# LEVERAGE MULTI-SOURCE TRAFFIC DEMAND DATA FUSION WITH TRANSFORMER MODEL FOR URBAN PARKING PREDICTION

YIN HUANG [a], YONGQI DONG [b*], YOUHUA TANG [a*], LI LI [c]

[a] *School of Transportation and Logistics, Southwest Jiaotong University, China*
Email: huangyin@my.swjtu.edu.cn; tyhctt@swjtu.cn

[b] *Institute of Highway Engineering, RWTH Aachen University, Aachen, Germany*
Email: yongqi.dong@rwth-aachen.de

[c] *Department of Automation, BNRist, Tsinghua University, China*
Email: li-li@tsinghua.edu.cn

**ABSTRACT**

The escalation in urban private car ownership has worsened the urban parking predicament, necessitating effective parking availability prediction for urban planning and management. However, the existing prediction methods suffer from low prediction accuracy with the lack of spatial-temporal correlation features related to parking volume, and neglect of flow patterns and correlations between similar parking lots within certain areas. To address these challenges, this study proposes a parking availability prediction framework integrating spatial-temporal deep learning with multi-source data fusion, encompassing traffic demand data from multiple sources (e.g., metro, bus, taxi services), and parking lot data. The framework is based on the Transformer as the spatial-temporal deep learning model and leverages K-means clustering to establish parking cluster zones, extracting and integrating traffic demand characteristics from various transportation modes (i.e., metro, bus, online ride-hailing, and taxi) connected to parking lots. Real-world empirical data was used to verify the effectiveness of the proposed method compared with different machine learning, deep learning, and traditional statistical models for predicting parking availability. Experimental results reveal that, with the proposed pipeline, the developed Transformer model outperforms other models in terms of various metrics, e.g., Mean Squared Error (MSE), Mean Absolute Error (MAE), and Mean Absolute Percentage Error (MAPE). By fusing multi-source demanding data with spatial-temporal deep learning techniques, this approach offers the potential to develop parking availability prediction systems that furnish more accurate and timely information to both drivers and urban planners, thereby fostering more efficient and sustainable urban mobility.

## 1. INTRODUCTION

With rapid urbanization and the significant increase in private vehicle ownership, the challenge of urban parking has emerged as a critical concern in metropolitan areas. Parking inaccessibility not only leads to longer driving times but also exacerbates urban traffic congestion, driver frustration, and environmental pollution (Bock et al., 2020). The challenge of finding parking spaces, especially during peak periods, is not solely due to the scarcity of parking spots but also largely attributed to the lack of predictive analytics on parking availability. Precise prediction of parking availability empowers city authorities and drivers to make informed decisions, optimize the utilization of parking resources, and effectively alleviate traffic congestion.

Methodologies for parking prediction can be broadly categorized into two groups: knowledge-based methods and data-driven approaches. The knowledge-based approach, reliant on prior knowledge and assumptions, is effective for simple, well-defined problems but inadequate for the complex task of parking availability prediction. Regarding data-driven parking prediction methods, three major phases have been experienced: statistical learning, machine learning (ML), and deep learning (DL). In short-term parking prediction, traditional statistical learning techniques like Kalman filters (Bock et al., 2020), Autoregressive Integrated Moving Average (ARIMA)(Bei Chen et al., 2013), and Historical Average (HA) (Chawathe, 2019) have demonstrated strong performance. However, for long-term forecasts, typical statistical approaches struggle to achieve high prediction accuracy due to temporal fluctuations and parking system complexity. To address the temporal variability and complex dynamics of parking, a number of studies turned to ML techniques. For example, Feng et al. (2019) employed ML algorithms, such as linear regression (LR), decision tree (DT), and random forest (RF), to analyze and predict parking availability, with the RF model demonstrating the highest accuracy. Some other studies have also compared the performance of algorithms such as support vector machine (SVM), K-nearest neighbors (KNN), and artificial neural networks (ANN) in parking prediction, e.g., in (Awan et al., 2020).

In recent years, DL techniques have emerged as prominent methods in parking prediction research. For instance, Qiu et al. (2018) combined recurrent neural networks (RNN) and genetic algorithms to develop an effective parking space forecasting model. To compensate for the spatial-temporal correlation between parking lots, Ghosal et al. (2019) developed a deep learning model that incorporates a convolutional neural network (CNN) and a stacked Long Short-Term Memory (LSTM) autoencoder. These aforementioned parking prediction researches concentrate on intricate model architecture, resulting in higher prediction accuracy but also consuming significant processing resources. A common oversight is neglecting the close correlation between parking availability and the attributes of the parking zone, as well as the dynamic traffic and trip demand in the associated area during the target time periods.

To address these issues, this study developed a spatial-temporal DL framework for parking availability prediction by fusing multi-source heterogeneous demanding data. Specifically, this study considered the spatial-temporal correlation among parking lots with similar characteristics within certain areas and constructed the "parking cluster zone" (PCZ) using the K-means clustering. Within the radiation area of the parking lot, travel records of metros, buses, online ride-hailing, and taxis related to the PCZ were integrated to build a set of associated demanding-related features. This study developed a spatial-temporal DL model based on Transformer with the attention mechanism, and compared it with various baseline models including traditional statistical models (e.g., HA and ARIMA), traditional ML (e.g., DT and SVR), ensemble ML (e.g., RF and XGBoost), and other DL models (e.g., GRU and LSTM). To test and evaluate the model performance, real-world empirical data from Chengdu, China were utilized for experiments. The results reveal that the proposed spatial-temporal DL framework outperforms state-of-the-art benchmarks in parking availability prediction.

In short, the main contributions of this paper lie in:
- For the first time, multi-source heterogeneous travel demanding data is aggregated and integrated to serve as the important spatial-temporal feature for parking availability prediction.
- A clustering-based method is adopted to establish parking cluster zones identifying parking lots with similar characteristics within certain urban areas. The spatial-temporal parking dynamics and correlations among related parking lots within the identified parking cluster zones are excavated.
- A spatial-temporal Transformer model is designed and customized. Together with various baseline models, the model performances are evaluated using real-world empirical data.

## 2. METHODOLOGY

### 2.1 Demanding Features Integration

#### 2.1.1 Parking cluster zone (PCZ)

The PCZ is defined as the area where travelers engage in various trip-related activities after parking their vehicles. The dynamic behavior of vehicles entering and exiting a parking lot is directly correlated with the demand characteristics of the surrounding area. To investigate parking availability more efficiently, this study utilized K-means clustering to group similar parking lots within certain areas together. As shown in **Fig. 1**, each "x" point represents a parking lot; different colors represent distinct parking lot clusters; "▲" signifies the cluster's center. Considering the layout of city streets, this study employs the Minkowski distance to compute the necessary buffer zone and then apply spatial concatenation to obtain the relevant PCZs. Four typical examples of PCZs are illustrated by the irregular polygons with boundary dashed lines framing relevant parking lots.

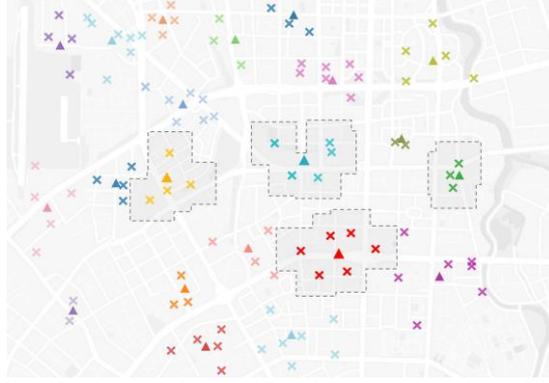

Figure 1. The distribution of the parking lots and parking cluster zone (PCZ)

With the advancement of comprehensive transportation integration and the widespread adoption of information and communications technologies (ICT), data-driven fusion research between heterogeneous modes of transportation has emerged as an effective approach to address the challenge of dynamic demanding estimation. Recognizing their potential interchangeability in offering travel services akin to private cars, this study selected four distinct travel modes for spatial-temporal demand feature fusion: metro, bus, online ride-hailing, and taxi. As illustrated in **Fig. 2**, upon extracting travel trip data from these four modes, the origin (O) and destination (D) of each trip can be identified. If either the origin (O) or destination (D) of a certain trip falls within the boundary of the PCZ, the multi-modal travel trip is then associated with that specific PCZ. In this study, the volume of such multi-modal travel trips is fused to demonstrate the spatial-temporal demanding characteristics of the PCZs. Consequently, this demand characteristic encapsulates both spatial and temporal aspects related to parking dynamics.

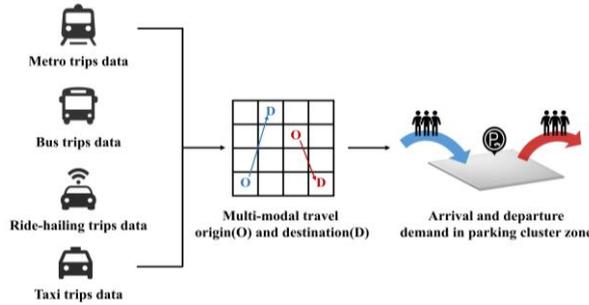

Figure 2. Fusion of demand characteristics within parking cluster zone (PCZ)

## 2.2 Transformer Model

Transformer, as illustrated in **Fig. 3**, is a DL model architecture based on attention mechanisms, with encoding and decoding components at its core (Vaswani et al., 2017). Each encoder and decoder consists of a multilayer self-attention mechanism and a feed-forward neural network. In comparison to CNN or RNN, the self-attention mechanism enables the Transformer model to prioritize the importance of different inputs, capturing long-range dependencies in the time series data more efficiently.

### 2.2.1 Positional Encoding

Positional encoding provides information about token positions by being added to the input embeddings before they are fed into the model, allowing Transformers to interpret relative positions and maintain sequential context. In time series forecasting tasks, temporal positional encoding represents converting timestamps into a format that can be effectively used by ML models. One common method is to use *sine* and *cosine* functions to encode time of day, day of the week, month, etc.

With "timestamp" as its associated timestamp and "T" representing the maximum value of the time feature (e.g., maximum hour in a day, maximum day in a week), the timestamp encoding can be computed as follows:

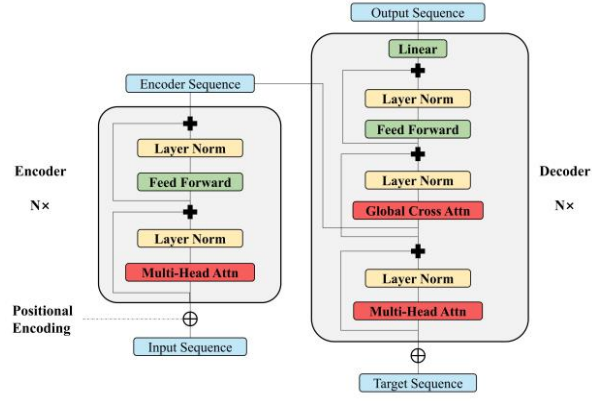

Figure 3. Transformer framework for time series forecasting

$$PE(timestamp, f)_{sin} = sin\left(timestamp \times \frac{2\pi}{T}\right) \quad (1)$$

$$PE(timestamp, f)_{cos} = cos\left(timestamp \times \frac{2\pi}{T}\right) \quad (2)$$

where $f \in \{hour, day, week, month\}$ represents the periodic feature of the input embeddings.

### 2.2.2 Multi-Head Attention

The attention mechanism enables the Transformer-based model to focus on various segments of the input sequence while generating an output sequence, thereby effectively capturing long-range dependencies. With Query-Key-Value (QKV) interior design, the scaled dot-product attention used in Transformer is given by

$$Attention(Q, K, V) = soft\,max\left(\frac{QK^T}{\sqrt{d_k}}\right)V \quad (3)$$

where $Q$, $K$, and $V$ are vector representations of the query, key, and value, respectively, and $d_k$ is the dimensionality of the key. $Q$, $K$, and $V$ are calculated by linearly converting the input sequence.

In multi-head attention, the query, key, and value matrices are divided into multiple heads, each with its own set of parameters. With $h$ attention heads, the multi-head attention mechanism is formulated as follows:

$$\text{MHAttn}(Q, K, V) = Concat(head_1, \cdots, head_H)W^O \quad (4)$$

where $head_i = Attention(QW_i^Q, KW_i^K, VW_i^V)$, and $W_i^Q, W_i^K, W_i^V$, and $W^O$ are learnable parameters.

### 2.2.3 Feed Forward Network

In the Transformer model, the Feedforward Neural Network (FFN) is connected to each attention layer. The feed-forward network is a fully connected module, defined as

$$\text{FFN}(H') = ReLU(H'W_1 + b_1)W_2 + b_2 \quad (5)$$

where $H'$ is outputs of the previous layer, $W_1 \in \mathbb{R}$, $W_2 \in \mathbb{R}$, $b_1 \in \mathbb{R}$, $b_2 \in \mathbb{R}$, are trainable parameters.

## 3. EXPERIMENTS AND RESULTS COMPARISON

### 3.1 Data Description and Processing

This study utilized parking records from 127 parking lots in Chengdu, Sichuan Province, China. For each parking record, one could obtain the vehicle's arrival time, departure time, date, and total number of parking spots in the relevant parking lot. The data collection spanned one month, from September 1 to September 30, 2021, and contained in total of 2,875,320 parking records during the 30 days. These 127 parking lots were clustered into 24 parking cluster zones (PCZs) based on factors such as spatial distribution and in/outflow. Additionally, travel trip records (including metro, bus, online ride-hailing, and taxi) within these 24 PCZs were also gathered. Note that, for practical limits (no card swipes are necessary to exit the bus service in the region), the bus OD trip demand data is based solely on boarding card swipe data.

For data preprocessing, this study initially employed the *dropna* function to discard instances containing *NULL* or missing values. Subsequently, the data was processed in sets with a 10-minute interval to further reduce noise. To mitigate the impact of low-frequency noise, a Fourier Transform was applied to the dataset, followed by an inverse Fourier Transform to revert the frequency domain data back to the time domain. After these treatments, 4,032 refined records were obtained from each parking lot, resulting in a total of 512,064 data samples (4,032*127 parking lots). The dataset was divided into two parts: 70% for training and 30% for testing. In the training dataset, data from 90 parking lots in 18 clusters were utilized for training. The training process involves using information from the past 432 time steps as input to predict the parking availability for the subsequent 144$^{th}$ time step. As each time step represents 10 minutes, the prediction for the subsequent 144$^{th}$ time step is equivalent to forecasting parking availability 24 hours (1 day) later, which is reasonable and holds practical significance in real-world applications. For the test dataset, data from 37 parking lots across 6 clusters were employed.

## 3.2 Evaluation Metric

To better demonstrate the performance of the models, this study incorporates the following evaluation indicators for parking availability prediction.

- Mean Squared Error (MSE):
$$\text{MSE} = \frac{1}{N} \sum_{n=1}^{N} (\hat{y}_n - y_n)^2 \quad (6)$$

- Mean Absolute Error (MAE):
$$\text{MAE} = \frac{1}{N} \sum_{n=1}^{N} |\hat{y}_n - y_n| \quad (7)$$

- Mean Absolute Percentage Error (MAPE):
$$\text{MAPE} = \frac{100\%}{N} \sum_{n=1}^{N} \left| \frac{\hat{y}_n - y_n}{y_n} \right| \quad (8)$$

Here, $N$ signifies the total number of test samples, while $y_n$ and $\hat{y}_n$ represent the actual and predicted values, respectively. MSE, MAE, and MAPE are three prevalent metrics for predictive analysis. Moreover, the model parameter size, represented as Params (M), along with the multiply-accumulate operations, denoted as MACs (G), serve as indicators of the DL models' complexity. The two metrics are frequently utilized to estimate models' computational complexity and real-time capabilities.

## 3.3 Results Comparison

To verify the effectiveness of the proposed spatial-temporal DL model based on Transformer, this study considers three types of baseline models, including traditional statistical models, i.e., HA and ARIMA), traditional ML, i.e., DT (Breiman et al., 1984) and SVR (Smola et al., 2004), ensemble ML, i.e., RF (Breiman et al., 2016) and XGBoost (Chen et al., 2016), and other DL models, i.e., GRU (Chung et al., 2014), LSTM (Hochreiter et al., 1997), and NLinear (Zeng et al., 2022). **Table 1** summarizes the quantitative comparison results for the evaluated models. It displays the mean values of the results across the target 37 parking lots. As demonstrated, the Transformer model outperformed other methods, with the lowest MSE, MAE, and MAPE scores. Conventional statistical models like HA and ARIMA and traditional ML models, e.g., DT and SVR, are inadequate in capturing complex spatial-temporal dependencies in parking availability. In comparison, XGBoost and RF algorithms notably improve prediction performance by leveraging larger ensembled model structures and iterative training. However, these ML algorithms do not outperform DL algorithms (i.e., GRU, LSTM, Transformer, NLinear). This may be because ML algorithms tend to overfit the training dataset, resulting in inferior performance on the test dataset. In contrast, DL models can automatically extract useful features and correlations, and they often employ regularization or penalty methods to mitigate overfitting.

To comprehensively compare the performance of different DL models, computational complexity and real-time capabilities are also taken into account. Among DL methods, traditional RNN models (GRU, LSTM) exhibit competitive performance in parking availability prediction. These algorithms capture temporal dependencies in sequences through gated units, but they also entail a larger number of parameters to be learned, which amounts to 1.8952 (M) for LSTM and 1.7781 (M) for GRU, respectively. Nevertheless, handling long-term sequence correlations remains a challenge for LSTM/GRU algorithms. This is further underscored by the findings of the ablation study depicted in **Fig. 4**.

On the other hand, the improved simple linear DL model, i.e., NLinear, is significantly smaller than LSTM and GRU in terms of MACs and Params. Despite its compact size, NLinear achieves superior performance through direct multi-step prediction. Among all DL models, the best-performing model, i.e., Transformer, demands the most computing resources with 1.0644 (G) MAC operations, yet only comprises 0.9107 (M) parameters. This is attributed to Transformer's self-attention mechanism, which reduces model parameters while amplifying processing complexity. Indeed, the significance of the self-attention mechanism lies in its ability to enable the Transformer model to efficiently utilize the spatial-temporal features, correlations, and dependencies within the integrated spatial-temporal demanding data and relevant parking data. By attending to relevant features across different time steps and PCZs, the Transformer model can capture complex relationships and dependencies, ultimately enhancing its predictive capability. Concerning this, it is worth noting that unlike in (Zeng et al., 2022), where NLinear outperformed Transformer, in this study, Transformer surpassed NLinear due to its efficient utilization of spatial-temporal features and correlations, while NLinear lacks in the spatial aspect.

Table 1. Performance Comparison of the Models

| Algorithms | MSE | MAE | MAPE | MACs(G) | Params (M) |
|---|---|---|---|---|---|
| HA | 0.2478 | 0.1748 | 1.1495 | --- | --- |
| ARIMA | 0.1761 | 0.1679 | 0.8785 | --- | --- |
| DT | 0.1477 | 0.1863 | 0.7723 | --- | --- |
| SVR | 0.1344 | 0.1637 | 0.7465 | --- | --- |
| XGBoost | 0.1162 | 0.1643 | 0.7275 | --- | --- |
| RF | 0.1017 | 0.1473 | 0.6704 | --- | --- |
| GRU | 0.0924 | 0.1665 | 0.6138 | 0.6876 | 1.7781 |
| LSTM | 0.0745 | 0.1582 | 0.6055 | 0.7216 | 1.8952 |
| NLinear | 0.0703 | 0.1542 | 0.5708 | **0.3185** | **0.6246** |
| Transformer | **0.0626** | **0.1358** | **0.5496** | 1.0644 | 0.9107 |

### 3.4 Ablation Study

To investigate the effect of the proposed integrated demand features on the model prediction results, ablation experiments were conducted regarding the utilization of features and regarding different prediction steps. There are four settings for the feature utilization: *1)* all the features, i.e., integrated spatial-temporal demanding data and the historical data of all the parking lots within the PCZs; *2)* integrated spatial-temporal demanding data plus only the historical data of the target parking lot; *3)* only the historical data of all the parking lots within the PCZs without integrated spatial-temporal demanding data; *4)* only the historical data of the target parking lot.

**Table 2** summarizes the prediction results of the top 2 models, i.e., NLinear and Transformer, under different feature settings. A comparison between feature settings 1 and 2 indicates that including historical information of similar parking lots within the PCZs enhances the accuracy of parking availability prediction for the target parking lot. Moreover, comparing the results of setting 1 versus 3 and setting 2 versus 4, highlights the significant improvement in prediction accuracy when incorporating the designed integrated spatial-temporal demand feature. Additionally, comparing the results of settings 3 versus 4 suggests that when using only the historical data from the parking lots, it is wise to utilize only the historical data of the target parking lot. Further examination is needed to understand why adding historical information from similar parking lots within the PCZ without adding the integrated demand feature worsens the prediction under this setting. Overall, the integration of multi-source heterogeneous demand data significantly contributes to parking availability forecasting.

To evaluate the model performance for different prediction steps, **Fig. 4** displays the prediction results of various algorithms validated on the test dataset with different time step settings. Traditional ensemble ML models (RF and XGboost), as well as classical RNN models (GRU and LSTM), suffer from error accumulation with increasing prediction time steps, leading to diminished performance in long-time series prediction. Conversely, the Transformer and NLinear models demonstrate proficiency in learning long-term dependencies, enabling them to perform well for both short and long-term step prediction.

Table 2. Prediction Results of Different Feature Settings

| Features settings | NLinear | | Transformer | |
|---|---|---|---|---|
| | MSE | MAE | MSE | MAE |
| 1 | 0.0713 | 0.1597 | **0.0626** | **0.1358** |
| 2 | 0.0764 | 0.1435 | 0.0832 | 0.1581 |
| 3 | 0.1808 | 0.2624 | 0.1697 | 0.2304 |
| 4 | 0.1647 | 0.2447 | 0.1452 | 0.1785 |

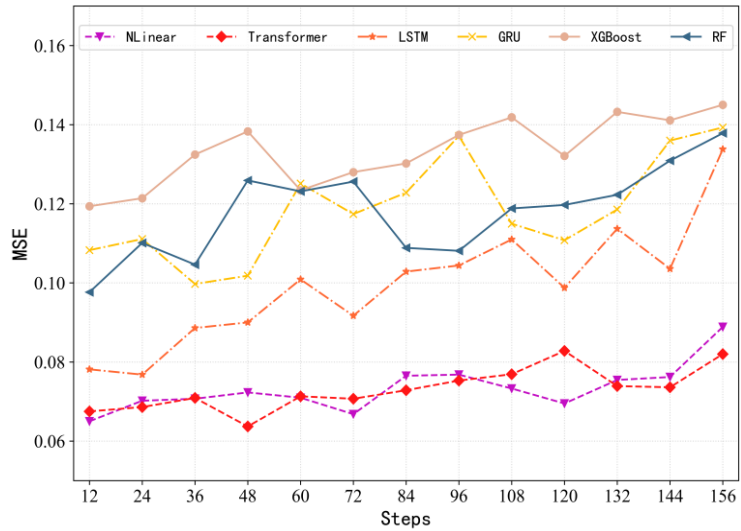

Figure 4. The results of MSE (Y-axis) for different prediction time steps (X-axis).

## 4. CONCLUSION

In the era of rapid urbanization, resolving parking issues is critical to alleviating urban traffic congestion and improving traffic operation efficiency. With the booming Internet of Things (IoT), Information and Communications Technology (ICT), and Artificial Intelligence (AI) technologies, it is possible to develop parking availability prediction systems. Although there are existing studies that have considered different machine learning (ML) and deep learning (DL) models for parking prediction, most of these studies are based on ineffective feature design with only ego parking lots data, which fails to capture effective spatial-temporal influencing factors. To address the gap and to improve the prediction accuracy of urban parking availability, this study designs a pipeline that integrates spatial-temporal Transformer-based DL with integrated spatial-temporal demanding data fused from multi-sources (i.e., metro, bus, online ride-hailing, and taxi services data, as well as parking lot data). The proposed method leverages K-means clustering to establish parking cluster zones (PCZs) grouping relevant parking lots together and extracting and integrating traffic demand characteristics from various transport modes within the target PCZ. Real-world empirical data was employed to verify the effectiveness of the proposed method.

Through extensive experiments, the performance of various traditional statistical models (HA, ARIMA), ML models (DT, SVR), ensemble learning models (RF, XGBoost), and DL models (GRU, LSTM, Nlinear, Transformer) on parking availability prediction is evaluated. The results demonstrate that DL algorithms significantly outperform traditional ML algorithms and statistical methods. More specifically, the Transformer-based model customized in this study achieved the highest performance, yielding the lowest Mean Squared Error (MSE), Mean Absolute Error (MAE), and Mean Absolute Percentage Error (MAPE). The ablation study regarding different feature utilization settings demonstrates the efficacy of the proposed integrated spatial-temporal demanding feature and the utilization of the designed PCZs to group similar parking lots together. Furthermore, the ablation study regarding different prediction steps verifies the proficiency of Transformer and NLinear models in learning long-term dependencies for delivering good performance in both short and long-term step prediction. This study can improve parking prediction systems, offering more accurate information to drivers and urban planners, thus enhancing urban mobility efficiency and sustainability.

## ACKNOWLEDGMENTS

This work was supported by the Science and Technology Program of Sichuan Province, China, Grant Number: 2021YFH0041, and the Natural Science Foundation of Beijing, China, Grant Number: L231007.